\documentclass[runningheads,a4paper]{llncs}
\usepackage{amsmath} 
\usepackage{amssymb}  

\usepackage{mathdots}
\usepackage{marvosym}
\usepackage{verbatim}
\usepackage{graphicx}
\usepackage{color}
\usepackage{url}
\usepackage{appendix}
\usepackage{cite}

\newtheorem{defn}[theorem]{Definition}

\newcommand{\keywords}[1]{\par\addvspace\baselineskip
\noindent\keywordname\enspace\ignorespaces#1}
\begin{document}

\mainmatter  

\title{Signal Classification using Smooth Coefficients of Multiple Wavelets to Achieve High Accuracy from Compressed Representation of Signal}

\titlerunning{Signal Classification via Wavelets}

%
%
\author{Paul Grant$^{\textrm{(\Letter)}}$ %
\and Md Zahidul Islam}
%
\authorrunning{ Paul Grant and Md Zahidul Islam}


\institute {School of Computing and Mathematics, Charles Sturt University \\
Bathurst, NSW 2795, Australia.\\ 
\email{\{pgrant,zislam\}@csu.edu.au}}


%
%

\toctitle{Lecture Notes in Computer Science}
\tocauthor{Authors' Instructions}
\maketitle

\begin{abstract}
Classification of time series signals has become an important construct and has many practical applications. With existing classifiers  we may be able to accurately classify signals,  however that accuracy may decline if  using a reduced number of attributes. 
Transforming the data then undertaking reduction in dimensionality may improve the quality of the data analysis, decrease time required for classification and simplify models.
We propose an approach, which  chooses suitable wavelets to transform the data,  then combines the output from these transforms to construct a dataset to then apply  ensemble classifiers to.
We demonstrate this on different data sets, across different classifiers and use differing evaluation methods.  Our experimental results demonstrate the effectiveness of the proposed technique, compared to the approaches that use either raw signal data or a single wavelet transform. 


\end{abstract}

\section{Introduction}
In this paper which we build upon our previous work \cite{gran:2019} where an approach using multi-wavelet decomposition to build a set of attributes composed of wavelet coefficients derived from different wavelet filters to enable noisy signal classification. Here  present a  method which first determines a manner to select suitable wavelets, then implements and combines these selected  wavelets to not only transform but reduce the dimension of the data and yet still maintain or improve upon classification accuracy when using decision trees as our classifier.\par
Use of decision tree induction for classification has become a  common approach in machine learning. Decision trees attempt to allot  symbolic decisions to new samples and provide a visual representation of the derived rule set. Automatic rule induction systems for inducing classification rules have already proved valuable as tools for assisting in knowledge acquisition for expert systems \cite{Innoise}.
  Classification is a methodology that determines categories within a  collection of data, which then  allows the analysis of large data sets. If we use labeled datasets to train algorithms that classify the data  it is considered an instance of supervised learning.
In most research, a signal is related to the main topic of interest. Here we apply wavelet transforms to that signal.
We utilise 
the smooth components from multiple wavelets to provide the attributes used for the classification process. 
We demonstrate Classification across 3 different sets of  signals\footnote{We designate these signals as raw or unmodified Data.}, where we apply classifiers  to both raw and transformed signals.  We show that accuracy may be improved or even maintained with reduced elements in the attribute space by using wavelets, and even further enhanced by multiple wavelet transformed data combined with ensemble of trees classifiers. 

Wavelet coefficients in preference to raw data has been previously mentioned \cite{WVoiceClass}, where a single wavelet filter was used. 
Using wavelet transformed data to select various frequency levels within the signal, enabling reduction or elimination of specific inherent frequencies (\textit{and noise}), to then undertake classification has proved successful \cite{WMusicClass}. 
 Wavelets may be used to reduce noise\footnote{Noise here includes missing or misclassification of values as well as other induced random fluctuations in the data.} in time series data, with the aim to provide better classification performance  which may be achieved after conducting a wavelet transform on the original data, also compression using wavelets speeds up  the classification process \cite{Dao2016}. Multiple Wavelets have been previously used \cite{Mort2005, gran:2019} and applied to noise reduction as well as Image and ECG data compression. However finding a close to optimal wavelet filter (\textit{or filters}) to use which provides a close to best representation the underlying data, is not a trivial task.\par

\section{Wavelets} \label{sub:Waves}
\vspace{-0.2cm}
Wavelets are linear transforms see Definition \ref{dfn:LinTran}, that can be used to segment the  data into separate non overlapping bandwidths.
 They have advantages where the signal has discontinuities and sharp spikes. Wavelets have been applied in various areas such as image compression, turbulence, human vision, radar and digital signal processing \cite{InWaves}. 
 A wavelet transform is the representation of a function by wavelet coefficients. 
\begin{defn}{Linear Transform}     \label{dfn:LinTran}

A linear transformation is a transformation $T : R^n \rightarrow R^m$ satisfying
\begin{eqnarray*}
T(u +v) & = & T(u) + T(v)\\
T(cu) & = & cT(u)
 \end{eqnarray*}
 for all vectors $u,v$ in $R^n$ and all scalars $c$.
 \end{defn}

 If we sample a wavelet discretely  and apply that resulting filter  applied to our raw data then we have a Discrete Wavelet Transform, (\textbf{DWT}). The DWT allows us to analyse \textit{(decompose)} a time series into coefficients \textbf{W}, from which we may synthesise \textit{(reconstruct)} our original series \cite{Perc2000}. 
The DWT in principle provides more information than the time series raw data points in classification because the DWT locates where the signal energies are concentrated in the frequency domain \cite{WVoiceClass}. We provide an overview of the DWT and the Multiple Discrete Wavelet transform (\textbf{MDWT}), adapted from \cite{gran:2019}. 
\subsubsection {DWT} Let a sequence $X_1, X_2,\ldots , X_n $ represent  a time series of $n$ elements, denoted as $\{X_t : t = 1,\ldots, n\}$ where $n= 2^J$ : $J \in \mathbb{Z^+},\; X_t  \in \mathbb{R}$, the discrete wavelet transform is a linear transform which decomposes $X_t$ into $J$ levels giving $n$ DWT coefficients;  the wavelet coefficients are obtained  by premultiplying $X$  by $\mathcal{W}$. 
\begin{equation}
 \textbf{W} = \mathcal{W}X
 \end{equation}
\begin{itemize}
    \item $\textbf{W}$ is a vector of DWT coefficients ($j$th component is $W_j$)
    \item $\mathcal{W}$ is $n \times n$ orthonormal transform matrix; i.e.,\newline
    $\mathcal{W}^T \mathcal{W} = I_n$, where $I_n$ is $n \times n$ identity matrix
    \item inverse of $\mathcal{W}$ is its transpose, $\implies \mathcal{W} \mathcal{W}^T = I_n$ \newline
    $\therefore \mathcal{W}^T$\textbf{W} = $\mathcal{W}^T \mathcal{W}X = X$
\end{itemize}
\textbf{W} is partitioned into $J + 1$ subvectors

\begin{equation}
  \textbf{W} =  [ \textbf{W}_1, \textbf{W}_2,\ldots, \textbf{W}_j, \ldots , \textbf{W}_J,\textbf{V}_J ]
\end{equation}


\begin{itemize}
    \item $\textbf{W}_j$ has $n/2^j$ elements \footnote{ note: $\sum^J_{j=1} \frac{n}{2^j} = \frac{n}{2} + \frac{n}{4} + \dots + 2 + 1 = 2^J -1 = n-1$}
    \item $\textbf{V}_J$ has one element \footnote{If we decompose $X_t: n=2^J$, to level $Jo : 1 \le J_0 \le J$ then $ \textbf{V}_{Jo}$  has $n/2^{Jo}$ elements }
\end{itemize}

conversely the synthesis equation for the  DWT is:
\begin{equation} \label{equ:syn}
 X =  \mathcal{W}^T \textbf{W} = \left[ \mathcal{W}^T_1, \mathcal{W}^T_2,\ldots, \mathcal{W}^T_J,  \mathcal{V}^T_J \right] \left[ \begin{tabular}{c} 
$\textbf{W}_1$ \\
$\textbf{W}_2$ \\
{\scriptsize$\vdots$}  \\

$\textbf{W}_J$ \\
$\textbf{V}_J$ \\
\end{tabular} \right]    
\end{equation}

 Equation \ref{equ:syn} leads to additive decomposition which expresses $X$ as the sum of $J+1$ vectors, each of which is associated with a particular scale $\mathcal{T}_j$

\begin{equation} \label{DWTadd}
X = \sum^J_{j=1} \mathcal{W}^T_j\textbf{W}_j + \mathcal{V}^T_J\textbf{V}_J  \equiv \sum^J_{j=1} \mathcal{D}_j + \mathcal{S}_J  
\end{equation}

\begin{itemize}
\item $\mathcal{D}_j \equiv \mathcal{W}^T_j\textbf{W}_j$ is portion of synthesis  due to scale $\mathcal{T}_j$, called the \textit{j}th 'detail'
\item $\mathcal{S}_J  \equiv \, \mathcal{V}^T_J\textbf{V}_j$ is a vector called the 'smooth' of the $J$th order \cite{Perc2000}.

\end{itemize}
\begin{remark}
 If we use the DWT to only decompose $X_t$ to the first level, then from Equation $\ref{DWTadd}$  the transformed signal consists of  $\frac{n}{2}$ detail coefficients and $\frac{n}{2}$ smooth coefficients. Similarly, decomposition to second level, $\frac{3n}{4}$ detail coefficients and $\frac{n}{4}$ smooth coefficients. \textit{(provided signal length n, is a factor of 2 or 4 respectively.)}
\end{remark}

\subsection{MDWT}  \label{subsec:MDWT1}To construct a Multiple Discrete Wavelet Transform from a time series $\{X_t : t = 1,\ldots, n\}$ we use the DWT to deconstruct the signal  to  level $J_0$ :  $J_0 \in \mathbb{Z^+}, \; X_t \in \mathbb{R}$, choose $N$ different DWT filters and apply to  $X_t$ sequentially. From Equation \ref{DWTadd} this results in:
\begin{equation} \label{eqn:MDWT}
 \sum^N_{i=1} \left[\sum^{J_0}_{j=1}  \mathcal{D}_{ij} + \mathcal{S}_{iJ_0} \right] 
\end{equation}

giving a sequence of vectors, where each $\text{DWT}_i$ is  $\mathcal{D}_{i1}, \mathcal{D}_{i2},\dots ,\mathcal{D}_{iJ_0},\;\mathcal{S}_{iJ_0}$
\textit{(which consists of the wavelet coefficients resulting from level $Jo$ decomposition).}  If we choose only the smooth coefficients $s_{i,k} \in \mathcal{S}_{iJ_0}: 1 \le k \le \frac{n}{2^{J_0}}$ then
\begin{equation}\label{eqn:smooth}
\textit{MDWT} = s_{1,1}, s_{1,2}, \dots , s_{1,\frac{n}{2^{Jo}}},  s_{2,1}, s_{2,2}, \dots , s_{2,\frac{n}{2^{Jo}}}, \ldots, s_{N,1}, s_{N,2}, \dots , s_{N,\frac{n}{2^{J_0}}}
\end{equation}
\begin{remark}
 This  has $\frac{N}{2^{J_0}}$ times as many elements as in $X_t$. 
\end{remark}

\subsection{Energy distribution} \label{subs:Energy}
If we define the energy with a signal $X_t$ as  the squared norm $||X||^2$, see Definition \ref{dfn:VectN}, then may derive the energy distribution in the signal  via a normalised partial energy sequence; $\mathbf{NPES}$ \cite{Perc2000}. 

 For a signal $\{X_t : t = 0,\dots, n-1\}$ if we reorder by magnitude then, 
 \begin{center}
   $|x_{(0)}| \ge |x_{(1)}| \ge \dots \ge |x_{(n-1)}|. $  
 \end{center}

This enables us to  compute the NPES\footnote{Which permits construction of a plot of cumulative energy\% in the  signal (or representation of), against the number of data points, see Figure \ref{NPESWAr}.
}, : $\;n-1 \ge M-1$

\begin{equation} \label{eqn:NPES}
 C_{M-1} \equiv \frac{\sum^{M-1}_{j=0}|x_{(j)}|^2} {\sum^{n-1}_{j=0}|x_{(j)}|^2} =
 \frac{\text{energy in largest M terms}}{\text{total energy in signal}}
\end{equation}

and similarly for a NPES of wavelet\footnote{As the DWT is an orthonormal transform, the energy in the transform \textit{(consisting of all $J + 1$  subvectors)} equates to the energy in the signal.} coefficients.
\newpage
\begin{defn}{Vector Norm} \label{dfn:VectN}

Given an $n$-dimensional vector $X = x_1, x_2, \dots, x_n$

The vector norm $||X||^p$ for $p = 1, 2, \ldots$ is defined as
\begin{equation*}
||X||^p \equiv \left\{ \sum_i |x_i|^p \right\}^{1/p}
\end{equation*}
\end{defn}

\section{Our Technique} \label{Sec:OurTECH}

We present a method,``Multi-Wavelet Compression Signal Classification", \newline
(\textbf{MWCSC}) which utilises wavelet transforms of the signal data, First we introduce the ideology of the main concepts followed by the advantages provided, then provide a succinct  overview of the steps required to implement our technique. \par
$\bullet$ From the dataset used, we consider the different classes within the signal data. Using wavelet transforms  of these classes, we map the distribution of the original signal energy/information against the corresponding representation provided by the wavelet coefficients.
We use the NPES, see section \ref{subs:Energy}, to select a group of suitable wavelet filters  to transform our data. NPES is an existing methodology that enables us to choose wavelets that may better represent the distribution of energy in the signal. We utilise the NPES with the aim providing the a sparse representation of the original data, i.e. many coefficients have low values.\par

$\bullet$ From multiple wavelets we construct a Multiple Discrete Wavelet transform, (MDWT, see section \ref{subsec:MDWT1}). The rationale for using multiple wavelets in the transform is, we may include wavelets that are either symmetric, have short support, provide higher accuracy and are orthogonal. No single wavelet may provide all such properties simultaneously \cite{Nason2008}.

\noindent
The MDWT provides the wavelet coefficients  used to form the attributes on which the classification methods derive their rule set from. For a transform consisting of a single wavelet we obtain the same number of data points (\textit{here wavelet coefficients}) as provided in the  original signal. However as we only choose the smooth wavelet coefficients $S_J$, see Equation \ref{DWTadd}, we may reduce the number of coefficients used to form the attributes for classification.\par

$\bullet$
At the first level of wavelet decomposition,  the MDWT consisting of $N$ wavelets (\textit{smooth coefficients only at $J=1$}), then from Equation \ref{eqn:MDWT} we would have have $\frac{N}{2}$ times the number of original data points, which could be considered as attributes. Similarly at the second level of decomposition, MDWT consisting of $N$ wavelets contains  $\frac{N}{4}$ time the number of original data points. \par
Similarly by reducing the overall number of attributes we may speed up the classification process \cite{Dao2016}. Our method enables us reduce the overall number of attributes yet still maintain or even improve classification accuracy. For evaluation of classification we utilise a different method for each dataset chosen, to highlight the results of this approach regardless of the manner used for grading the results.

\subsection{Advantages} \label{Advan}
Our methodology  does require some additional computation, however
\renewcommand{\labelitemi}{$\bullet$}
\begin{itemize}
\item The NPES identifies suitable wavelets to use,
\item By using only smooth coefficients from the wavelet transform, it is possible reduce the number of attributes for the classifiers
\item Using multiple wavelets increases accuracy

\end{itemize}
While the  construction of the MDWT is not  a complicated procedure,
the results returned  via the MDWT constructed using only the smooth coefficients highlight that the extra computation is worthwhile, providing us with similar levels of accuracy yet reducing number of attributes.

We are providing a set of attributes for the classifiers where a considerable amount of energy in the signal is then concentrated into a smaller of number of components. The classification methods  combined with MDWT tend to return smaller sets of decision rules (or smaller less complex trees) to arrive at their final rule set\cite{gran:2019}. 
Using our MDWT compared to a single wavelet transform,  in Sections \ref{sub:dataCl}  to \ref{subsec:FordA} the gain in classification accuracy when used with single decision tree  methods and even more so when used with ensemble classification methods, is apparent .
The software used in construction and application is freely available on-line together : R \cite{CRAN}, the R package WMTSA \cite{Wmtsa} and WEKA \cite{WEKA}. 

\subsection{Steps in the Proposed Technique: MWCSC} \label{MainMeth}
 \begin{itemize}
\item [] \textbf{Step 1 \textit{Wavelet Selection}}.
\item[]From a set of Time series  
 \[ \{X_{t_i} : t= 1, \ldots, n, \, i = 1, \ldots, K  \} \] 

 \item[]which has a distinct number of classes, compare  the energy distribution  of each of the signal classes as represented by varied wavelet transforms,  using the NPES described in section \ref{subs:Energy}. 
 \end{itemize}
 
\textbf{Step 2 \textit{Discrete Wavelet Transforms}}
\vspace{-0.2cm}
\begin{itemize} \label{MainProc}
\item []For each single time series $X_{t_i}$ take the DWT of the signal using a different wavelet filter (as determined by the NPES in \textbf{\textit{Step 1}}) for each of the transforms and extract the  wavelet smooth coefficients $\mathcal{S}_{J_0}$.

\end{itemize}

\textbf{Step 3 \textit{DataSet Construction via MDWT}}
\begin{itemize}

\item[]\textbf{3a}. Construct  a  new  data  series, (\textit{MDWT, see section \ref{subsec:MDWT1}}) placing  each  of  the  individual  vectors  of  the
wavelet smooth coefficients (resulting from each DWT, with level of decomposition = $J_0$) in a continuous sequence, one
after each other, see equation \ref{eqn:smooth}. 
\item[]\textbf{3b}. For each each of these MDWT, stack each transformed signal, to form a data
array or matrix as depicted in Table \ref{tab:MDWTt}.
\end{itemize}

 \begin{table}
\caption{ Array of transformed data as developed by MDWT}
\resizebox{\textwidth}{!}{%
\begin{tabular}{ccccccccccccc}\label{tab:MDWTt}
MDWT($X_{t_1}$): & $s_{1,1_1}$, & $s_{1,2_1}$, & \dots, & $s_{1,\frac{n}{2^{J_0}}_1}$, & $s_{2,1_1}$,  & \dots, & $s_{2,\frac{n}{2^{J_0}}_1}$,&\dots, & $s_{N,1_1}$,  & \dots, & $s_{N, \frac{n}{2^{J_0}}_1}$\\
MDWT($X_{t_2}$): & $s_{1,1_2}$, & $s_{1,2_2}$, & \dots, & $s_{1, \frac{n}{2^{J_0}}_2}$, & $s_{2,1_2}$, & \dots, & $s_{2, \frac{n}{2J_0}_2}$,&\dots, & $s_{N,1_2}$, & \dots, & $s_{N, \frac{n}{2^{J_0}}_2}$\\
& & & $\vdots$ &  & &  $\vdots$ &   & $\vdots$&  & $\vdots$  &\\
MDWT($X_{t_K}$): & $s_{1,1_K}$, & $s_{1,2_K}$, & \dots, & $s_{1, \frac{n}{2^{J_0}}_K}$, & $s_{2,1_K}$, & \dots, & $s_{2, \frac{n}{2^{J_0}}_K}$,&\dots, & $s_{N,1_K}$,  & \dots, & $s_{N, \frac{n}{2^{J_0}}_K}$
\end{tabular}}
\vspace{-0.4cm}
\end{table}

\textbf{Step 4 \textit{Build a classifier}}
\begin{itemize}
\item[] Using the new data as generated in previous steps, apply ensemble classifiers.\footnote{We also applied single Decision tree classifiers to the MDWT data to  as a baseline to compare with ensemble classifiers.} to the transformed data.
\end{itemize}

\section{Experimental Results} \label{sec:Exp}
The data used was sourced from the UCR time series archive \cite{UCRData}.
Here we simply chose three data sets, each of different length and number of records, see Table \ref{table:data}. These data sets exhibit widely different levels of smoothness\footnote{Smoothness defined here as: standard deviation of the of  first differences of a time series elements. ie. standard deviation of ($X_S$) : $X_S = x_1 - x_2, x_2 - x_3, \ldots, x_{n-1} - x_n $ } when compared to each other.  We also use 5 Tree based classifiers from WEKA.

\subsection{Classification methods used }
\label{sub:classM}
We utilise the following classifiers\footnote{ For the Ensemble Classifiers* throughout our experiment we set number of trees used to 100, no fine tuning of parameters was undertaken.}.
\begin{itemize}
\item \textbf{J48} a decision tree is an extension of ID3. Some additional features of J48; accounting for missing values, decision trees pruning, continuous attribute value ranges and derivation of rules \cite{Quin93}.
\item \textbf{Random Forest}* Class for constructing a forest of random trees.\cite{Brei2001}.

\item \textbf{ForestPA}* Decision forest algorithm Forest PA, 
using bootstrap samples and penalised attributes \cite{Adnan2017}. 
\item \textbf{SysFor}* Decision forest algorithm SysFor, a systematically developed forest of multiple decision trees \cite{Islam2011}.

\item \textbf{SimpleCart} Classification and Regression Tree, Class implementing minimal cost-complexity pruning \cite{Brei2017}.
\end{itemize}

\begin{table}[ht]
\caption{Dataset descriptions}
\vspace{-0.3cm}
\label{table:data}
\begin{center}
\begin{tabular}{ l c|c|c|c|} 
  \textit{Data Name} & Test Set  \; & Training Set \;  & No.of Classes  & Length \\
\cline{1-5}
 \textit{ArrowHead} & 35 & 176 & 3 & 251 \\
\cline{2-5}
\textit{Mallet} & 2345 & 55 & 8 & 1024\\
\cline{2-5}
\textit{FordA} & 1320 & 3601 & 2 & 500 \\
\cline{2-5}
\end{tabular}
\end{center}
\vspace{-1cm}
\end{table}


\subsection{Arrowhead data} \label{sub:dataCl}
\begin{figure}[ht]
     \centering
     \includegraphics[width=\textwidth, height = 5.8cm]{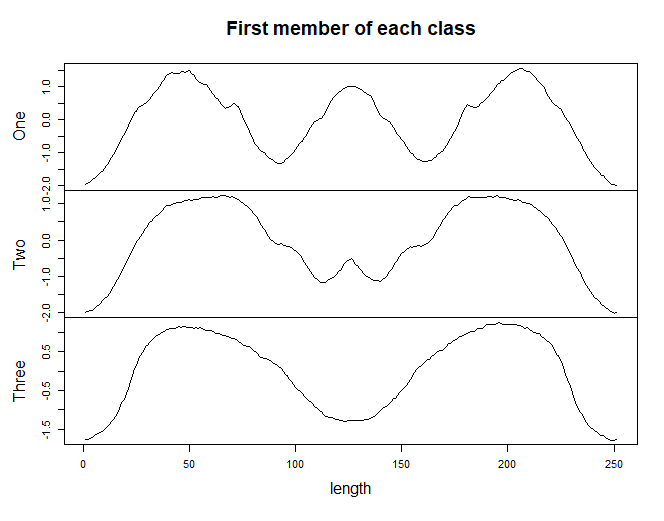}

      \caption{Profile of each Arrowhead class}
    \label{SignAr}
    \vspace{-0.3cm}
   \end{figure}

The first dataset consists of profiles of Arrowheads where we utilise the profiles as time series. The dataset has three classes, see Figure \ref{SignAr},  we combined the test and training sets together to give 211 records This enabled us to use \textit{Ten-fold Cross-Validation} with WEKA for evaluation.

      \begin{figure}[ht!]
     \centering
     \includegraphics[width=\textwidth, height = 6.6cm]{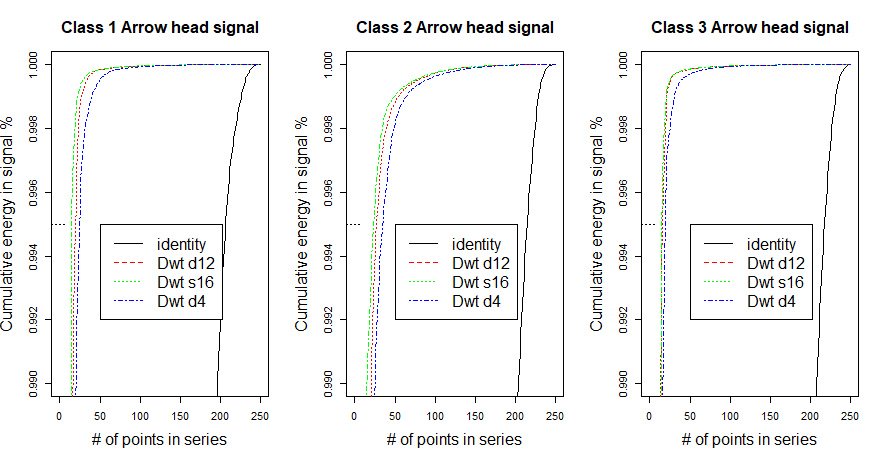}

      \caption{NPES of each Arrowhead Class using Wavelet Transforms}
    \label{NPESWAr}
    \vspace{-0.3cm}
   \end{figure}

   Following the $\mathbf{MWSCS}$ procedure, initially we applied various wavelet filters to each of the arrowhead classes to plot the NPES. Given the 251 data points in length, the wavelet transform deconstructing the signal to the first level, provided us with 125 detail and  125 smooth coefficients but also an ``Extra" class of coefficients, how this class is calculated and implemented in the transform, by the chosen software is fully described in \cite[Chap~4.11]{Perc2000}. \par
   These additional coefficients are simply untransformed data. For this dataset as the ``Extra" coefficients contain considerable signal energy/information, they are combined with the smooth coefficients in constructing our new data. For each transformed  signal, the Extra coefficients contain an average of  1.41\% of signal energy per transformed signal at level 1 and 3.45\% at level 2.
   
   The NPES of samples taken from each Arrowhead class highlight suitable wavelet filters to represent the energy in a smaller number of data points as shown in Figure \ref{NPESWAr}. Little difference is apparent between the wavelet filters shown, however the filters s16  and d12 represent the energy slightly  more efficiently than d4.
   Hence we use these wavelet filters to transform our signal data and build our transformed dataset, this however provides no compression in the signal representation as we will still have same number of  wavelet coefficients as data points  in the original signal. 
   \par
   By selecting only the smooth coefficients\footnote{ In this instance we also include the Extra coefficients as they  represent considerable signal energy.} at the respective level we are able to reduce the number of attributes within our data and maintain classification accuracy as shown in Table \ref{tableArrow}. Here classification accuracy is defined as the number of correctly Classified Instances with respect to the dataset's Class labels which are initially provided within the original data.\footnote{Using Accuracy \% here is a suitable metric, as the dataset is reasonably balanced i.e. Class 1 has 81 records, Class 2 and 3 have 65 records each.}
   
   \subsubsection{Arrowhead Results}
Table \ref{tableArrow} demonstrates the benefits of the wavelet transforms, s16 wavelet filter performs slightly better than d4\footnote{As indicated by the NPES, Figure \ref{NPESWAr}.} for the ensemble classifiers,  The use of the smooth coefficients resulting from s16 (\textit{and when combined with d12}), maintain accuracy while using reduced numbers of attributes.

Using the smooth wavelet coefficients and including the Extra coefficients, where these Extra coefficients contain considerable signal energy, still provide considerable accuracy, especially when using the ensemble classifiers. 

\begin{itemize}
 \item [] $\mathbf{Table \;\ref{tableArrow}\; Description:}$
\item $\mathbf{Classifier}$ The first column is the list of classifiers used, see section \ref{sub:classM}
\item $\mathbf{Raw \; data}$ results from classifying original data, being 251 units in length
\item $\mathbf{Wavelet\; d4}$ results from data transformed using Wavelet filter d4, signal decomposed to four levels, 
248 units in length
\item $\mathbf{Wavelet \; s16}$ results from data transformed using using Wavelet filter s16, signal decomposed to four levels, 
248 units in length
\item $\mathbf{Wavelet \; s16\; \mathcal{S}_1 + Extra}$  results from  data transformed using s16 using only the smooth level 1  and Extra coefficients, 126 units in length
\item $\mathbf{Wavelet\; s16\; d12\; combined\;\mathcal{S}_2}$  results from data transformed using s16  and d12 filters, combining  the smooth level 2 coefficients from both transforms to form a new data series , 62 + 62 = 124 units in length. A  MDWT, see section \ref{subsec:MDWT1}

\item $\mathbf{Wavelet\; s16\; d12 \;combined \;\mathcal{S}_2 + Extras}$  results from data transformed using s16  and d12 wavelet filters, combining  the smooth level 2  and Extra coefficients from both transforms to form a new data series , 64 + 64 = 128 units in length, again a  MDWT.

\end{itemize}

\begin{table}[h!]
\caption{Classification of Arrowhead data,  \textit{Cross Validation 10-Fold}}
\vspace{-0.8cm}
\label{tableArrow}
\begin{center}
\resizebox{\textwidth}{!}{%
\begin{tabular}{ c c |c|c|c|c|c|} 
 &\multicolumn{5}{c}{Classification Accuracy\%} \\
  \textit{} & Raw data \; & Wavelet d4 & Wavelet s16  & Wavelet s16  &Waves s16 d12 & Waves s16 d12   \\
  \textit{$\mathbf{Classifier}$} &  & 4 levels  & 4 levels & $\mathcal{S}_1$ + Extra  &combined $\mathcal{S}_2$ & comb. $\mathcal{S}_2$ + Extra\\
\cline{2-7}
 \textit{J48} & 75.35 &76.03  & 74.41& 79.14  & 77.72& 72.72\\
\cline{2-7}
\textit{Rforest} & 86.25 & 84.36 & 86.25& 90.05& 89.1 & 89.57\\
\cline{2-7}
\textit{ForestPA} & 80.09 & 79.14 & 81.51& 79.62 & 82.46& 83.88\\
\cline{2-7}
\textit{SysFor} & 82.46 & 81.52 & 81.99& 83.41 &84.36 & 84.36\\
\cline{2-7}
 \textit{SimpleCart} & 73.46 & 73.93 &  72.51& 70.14 & 70.61& 70.61\\
\cline{2-7}

\end{tabular}}
\end{center}
\end{table}
\subsection{Mallat data}
\begin{figure}[ht]
     \centering
     \includegraphics[width=\textwidth, height = 5.7cm]{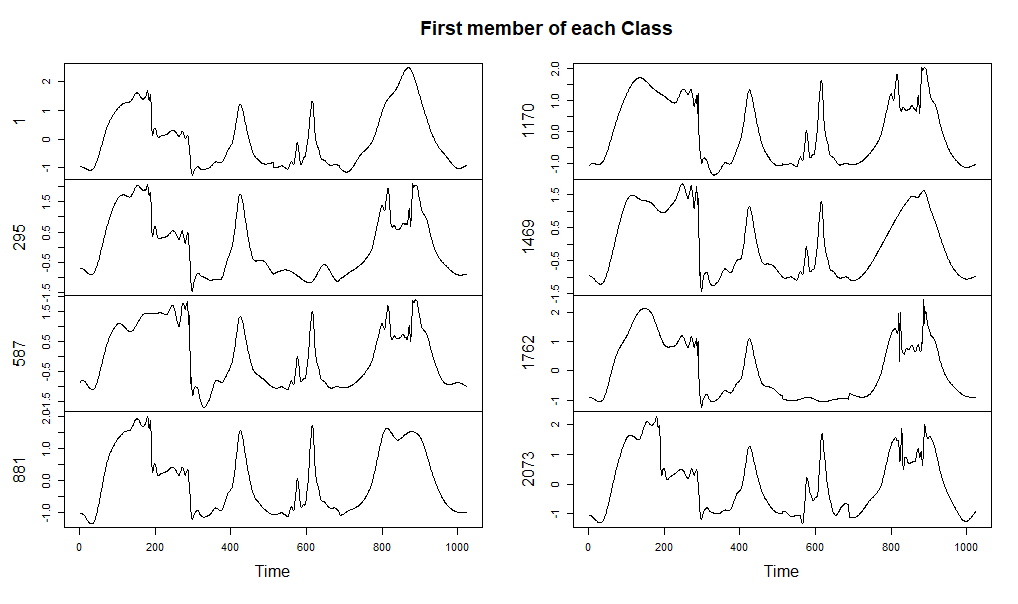}

      \caption{Mallat data}
    \label{MallatD}
    \vspace{-0.1cm}
   \end{figure}

From the UCR dataset, Mallat Curve data, 8 distinct classes of 1024 units
 in length, see Figure \ref{MallatD}. We combined the data, both testing and training sets to have a larger set containing 2400 records, \textit{eight classes of 300 records each}.

   Following the $\mathbf{MWCSC}$ procedure as with the Arrowhead data,  using the NPES to determine suitable wavelets to represent the signal energy, we apply the classifiers to the wavelet transformed data. This time we use 20\% of data for training and 80\% for testing, (the actual records in these sets were chosen by  WEKA).
   \subsubsection{Mallat data, Results} Table \ref{tableMallat} highlights results from the Classifiers. The original data is 1024 unit long, hence the full wavelet transform also 1024 units long. Using the smooth wavelet coefficients from $\mathcal{S}_1$ results 512 units in length, similarly $\mathcal{S}_2$ results in 256 units and $\mathcal{S}_3$, 128 units. \par
Using the various levels of decomposition, increasing the effective compression of the signal transform we  note that accuracy is maintained, especially with the ensemble classifiers.
However when we combine  the two  $\mathcal{S}_3$ levels (a MDWT) from different wavelets (as determined by the NPES), then the ensemble classifiers maintain considerable accuracy, given the reduced transformed signal size, $\mathcal{S}_3\times$2 = 128 + 128 = 256 units. A small accuracy gain over the single wavelet transforms  using $\mathcal{S}_2$ or $\mathcal{S}_3$.

   \vspace{-0.2cm}
   \begin{table}[ht]
\caption{Classification of Mallat data, 20\% training,  80\% testing}
\vspace{-0.9cm}
\label{tableMallat}

\begin{center}
\resizebox{\textwidth}{!}{%
\begin{tabular}{ c c | c|c|c|c|c|} 
 &\multicolumn{5}{c}{Classification Accuracy\%} \\
  \textit{} & Raw data \; & Wavelet s16 & Wavelet s16  & Wavelet s16  &Wavelet s16  & Waves s16 d8 \;  \\
  \textit{$\mathbf{Classifier}$} &  & 10 levels  & $\mathcal{S}_1$ & $\mathcal{S}_2$  & $\mathcal{S}_3$ & combined S3 \\
\cline{2-7}
 \textit{J48} & 98.88 &94.48  & 97.29 & 95.99  & 95.67& 94.74\\
\cline{2-7}
\textit{Rforest} & 97.18 & 98.85 & 98.33& 98.01& 98.07 & 98.25\\
\cline{2-7}
\textit{ForestPA} & 96.61 & 97.34 & 97.39& 97.23 & 97.81& 97.86\\
\cline{2-7}
\textit{SysFor} & 95.52 & 95.93 & 96.15& 94.63 &95.15 & 95.36\\
\cline{2-7}
 \textit{SimpleCart} & 95.57 & 95.0 &  96.3& 95.93 & 94.17& 94.53\\
\cline{2-7}

\end{tabular}}
\vspace{-0.5cm}
\end{center}
\end{table}


\subsection {Ford data} \label{subsec:FordA}
From the UCR dataset, Ford data, 2 distinct classes of 500 units in length. here the classification problem is to diagnose whether a specific symption exists or not in an automotive subsystem.  We use the  original training and test sets as provided in the UCI data and follow the $\mathbf{MWCSC}$. The test set, had 681 records in one class and 629 in the other. \par NPES determined  wavelet filters d16 and s20 as suitable.
Results in Table \ref{tableFord} show we may maintain accuracy even at higher levels of compression,(or attribute reduction).  Using the smooth components at level 1, $\mathcal{S}_1$ results in 250 units, using $\mathcal{S}_2$ provides 125 units,  $\mathcal{S}_3$ gives 62 units and $\mathcal{S}_3$ + Extra coefficient gives 63 units in length. Very little gain is evident in this last transform as only a minor levels of signal energy is contained within the Extra coefficients at the $3^{rd}$ level.\footnote{The average energy/information provided by the Extra coefficients at level $\mathcal{S}_3$, per transformed signal is only 0.5\%  hence little if at all any gain in accuracy is achieved by including the Extra coefficients at this level.}


   \begin{table}[h!]
\caption{Classification of Ford data, 3601 training records,  1320 testing records}
\vspace{-.7cm}
\label{tableFord}
\begin{center}
\resizebox{\textwidth}{!}{%
\begin{tabular}{ c c |c|c|c|c|c|} 
 &\multicolumn{5}{c}{Classification Accuracy\%} \\
  \textit{} & Raw data \; & Wavelet d16 & Wavelet d16  & Wavelet d16  &Waves d16  s20 & Waves d16 s20 \;  \\
  \textit{\textbf{Classifier}} &  & S1  & S2 & S3  &  combined S3 & comb. S3 + Extra\\
\cline{2-7}
 \textit{J48} & 56.13 &58.03  & 56.44 & 52.42  & 58.71& 58.18\\
\cline{2-7}
\textit{Rforest} & 73.25 & 72.95 & 71.37& 73.41& 74.2 & 75.53\\
\cline{2-7}
\textit{ForestPA} & 74.24 & 75.91 & 74.17& 74.69 & 74.09& 74.92\\
\cline{2-7}
\textit{SysFor} & 62.27 & 59.17 & 59.59& 63.79 &64.47& 60.38\\
\cline{2-7}
 \textit{SimpleCart} & 58.18 & 58.11 &  56.37& 59.02 & 59.92& 59.15\\
\cline{2-7}
\end{tabular}}
\end{center}
\end{table}
\begin{remark}
For the single tree method J48, It would appear that our choice of specific wavelets within the MDWT might not be crucial as no evidence of consistent additional gain between the two MDWT variations. However the  choice of additional attributes (as provided by the  MDWT) to train upon would seem to provide some  minor gain over using a single Wavelet transform. 
\end{remark}

\section{Conclusion}  

It has been shown previously that wavelets may be used for signal compression with little loss of accuracy \cite{Dao2016}.  Our MWCSC methodology which makes use of the NPES and MDWT provides us a method to determine suitable wavelets as well as add additional information to the attribute space. This enables us to use transformed data sets with smaller dimension than the original data yet still provide similar or enhanced accuracy.   From the NPES graphic, Figure \ref{NPESWAr}, we note that wavelet d4 is not as efficient at energy representation as wavelet s16, for the arrowhead dataset. This is similarly represented in Table \ref{tableArrow} by comparing classification results, across the various classifiers  of raw data as well as  the data transformed by the d4 or the  s16 wavelet filters. \par Construction of the MDWT from suitable wavelets enhances the accuracy while offering an effective compressed representation of the signal to which the classifiers my be applied to.   
Inclusion of ``Extra" coefficients into the construction  of the MDWT, where such coefficients contain considerable signal energy adds additional accuracy  for little extra computation,
as they are an included class in the transform calculation where $\{X_t : t = 1,\ldots, n\}, \; n \ne 2^J \; J \in \mathbb{Z^+}.$






\subsection{Future Work}
 Within  the construction of a MDWT, each  separate DWT used to decompose the time series  had the level or scale $J_0$ of decomposition remain unchanged. This may not be optimal as wavelet family, filter length and scale of decomposition, may all be correlated  with respect to how the energy/information in the signal is represented. Hence wavelet decomposition using different scales within a MDWT should be considered. The the number of wavelet filters within a MDWT might be altered depending upon importance of accuracy or compression (\textit{reduced number of attributes}), in our experiment  a MDWT consisted of only two different wavelet filters.

\end{document}